\documentclass[conference,letterpaper, 10pt]{ieeeconf}
\pdfoutput=1
\IEEEoverridecommandlockouts

\usepackage{times}

\usepackage[backend=biber,style=ieee,natbib=true,doi=false,isbn=false,url=false,mincitenames=1]{biblatex} 
\usepackage{multicol}
\usepackage[bookmarks=true]{hyperref}
\usepackage[table,xcdraw]{xcolor} %
\usepackage{ragged2e} %

\usepackage{amsmath,amsfonts,amssymb, bm}
\usepackage[utf8]{inputenc}
\usepackage{graphicx}
\usepackage{subcaption}
\usepackage{hyperref}
\usepackage[capitalise]{cleveref}
\usepackage{caption}
\usepackage{lipsum}%
\usepackage{array}
\usepackage{color}

\newcommand{\gtlocations}{\boldsymbol{G^\#}}
\newcommand{\gtsensedlocations}{\boldsymbol{X^\#}}

\newcommand{\sensedlocations}{\boldsymbol{X}}
\newcommand{\sensedvalues}{\boldsymbol{Y}}
\newcommand{\location}{g}
\newcommand{\sensedlocation}{x}
\newcommand{\sensedvalue}{y}
\newcommand{\sensingfunction}{\psi}

\newcommand{\objectivefunction}{f}
\newcommand{\argmax}{\arg\!\max}

\renewcommand{\t}[1]{{\textrm{#1}}}

\newcommand{\eg}{e.g.,~}
\newcommand{\etc}{etc.}

\newif\ifshowcomments

\ifshowcomments
    \newcommand{\gaurav}[1]{{\color{red} \textrm{Gaurav:} #1}}
    \newcommand{\todo}[1]{{\color{red} \textrm{TODO:} #1}}
    \newcommand{\chris}[1]{{\color{green} \textrm{CED: #1}}}
    \newcommand{\gautam}[1]{{\color{blue} \textrm{GSal: #1}}}
    
\else
    \newcommand{\gaurav}[1]{}
    
    \newcommand{\todo}[1]{}
    \newcommand{\chris}[1]{}
    \newcommand{\gautam}[1]{}
\fi

\usepackage{subfiles} %

\begin{document}
\title{Learned Parameter Selection for Robotic Information Gathering \\
\thanks{This work was supported in part by the Southern California Coastal Water Research Project Authority under prime funding from the California State Water Resources Control Board on agreement number 19-003-150 and in part by USDA/NIFA award 2017-67007-26154.}
\thanks{CED, GS, DC, and GSS are with the University of Southern California. {\tt \{cdennist,salhotra,dcaron,gaurav\}@usc.edu.} AK was an intern at USC when this work was done. He is currently at the University of Michigan {\tt \{akanga\}@umich.edu.} 
GSS holds concurrent appointments as a Professor at USC and as an Amazon Scholar. This paper describes work performed at USC and is not associated with Amazon.}%
}

\author{Christopher E. Denniston \and Gautam Salhotra \and 
Akseli Kangaslahti \and David A. Caron\and Gaurav S. Sukhatme}%
\maketitle

\begin{abstract}
When robots are deployed in the field for environmental monitoring they typically execute pre-programmed motions, such as lawnmower paths, instead of adaptive methods, such as informative path planning. One reason for this is that adaptive methods are dependent on parameter choices that are both critical to set correctly and difficult for the non-specialist to choose. Here, we show how to automatically configure a planner for informative path planning by training a reinforcement learning agent to select planner parameters at each iteration of informative path planning. We demonstrate our method with 37 instances of 3 distinct environments, and compare it against pure (end-to-end) reinforcement learning techniques, as well as approaches that do not use a learned model to change the planner parameters. Our method shows a 9.53\% mean improvement in the cumulative reward across diverse environments when compared to end-to-end learning based methods; we also demonstrate via a field experiment how it can be readily used to facilitate high performance deployment of an information gathering robot. 

\end{abstract}

\section{Introduction}

Robots are routinely deployed in aquatic environments for science~\cite{fossum_information-driven_2018} and environmental monitoring~\cite{hitz_ecological, seth_jfr, lawrance2019shared_seth_jfr}. Most applications in such settings involve taking measurements to better understand and model continuous processes such as ocean fronts~\cite{seth_jfr,joao_iser}, marine biochemistry~\cite{plume_tracking} and microbial ecology~\cite{das_sampling, robbins2006improved}. The dominant paradigm in these field deployments is for robots  to execute pre-planned paths, such as lawnmower trajectories~\cite{mbari_video}. 
On the other hand, adaptive sampling approaches~\cite{Hollinger2014, denniston_icra_2021}, in which the robot can make autonomous decisions on the fly, allow for greater flexibility and efficiency~\cite{joao_iser}. 
However, such approaches are rarely used in practice because they typically require significant parameter tuning and their behavior is difficult to predict and understand - formidable barriers to adoption by non-technical users. We focus on the first, parameter tuning,  for a broad class of adaptive sampling techniques collectively called informative path planning (IPP)~\cite{Hollinger2014}, in which a robot creates a plan to take sequential measurements to build an approximate map of some continuous process of interest by maximizing an objective function. The map is used to guide planning for further measurement or scientific study.

IPP-based systems have seen limited field use~\cite{fossum_information-driven_2018,seth_jfr} but their uptake is hampered by the need to specify complicated parameters. These parameters are abstract, difficult to explain to non-technical users, and can greatly affect system performance. In the worst case, mis-specifying them significantly reduces performance~\cite{seth_jfr, denniston_icra_2021} compared to a pre-planned path - a serious problem since a single missed day of data collection at sea can cost tens of thousands of USD on research vessels~\cite{NAP12775}.
Conversely, correctly specifying them can more than double information gathering performance in some scenarios~\cite{denniston_icra_2021}.

\begin{figure}[t]
    \begin{subfigure}{.67\columnwidth}
        \centering
        \includegraphics[width=\textwidth]{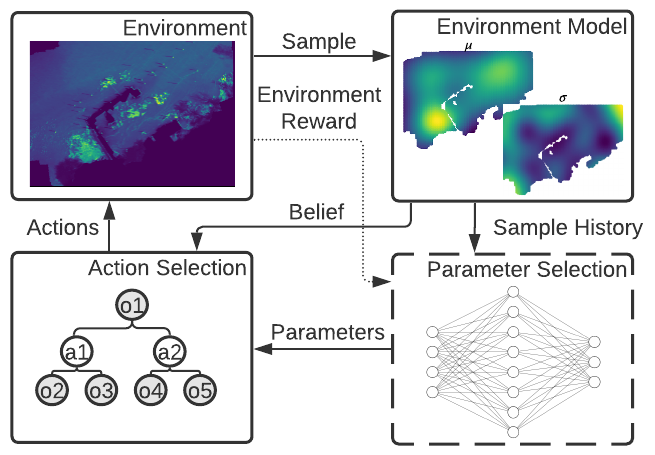}
    \end{subfigure}
    \begin{subfigure}{.32\columnwidth}
        \centering
        \includegraphics[width=\textwidth]{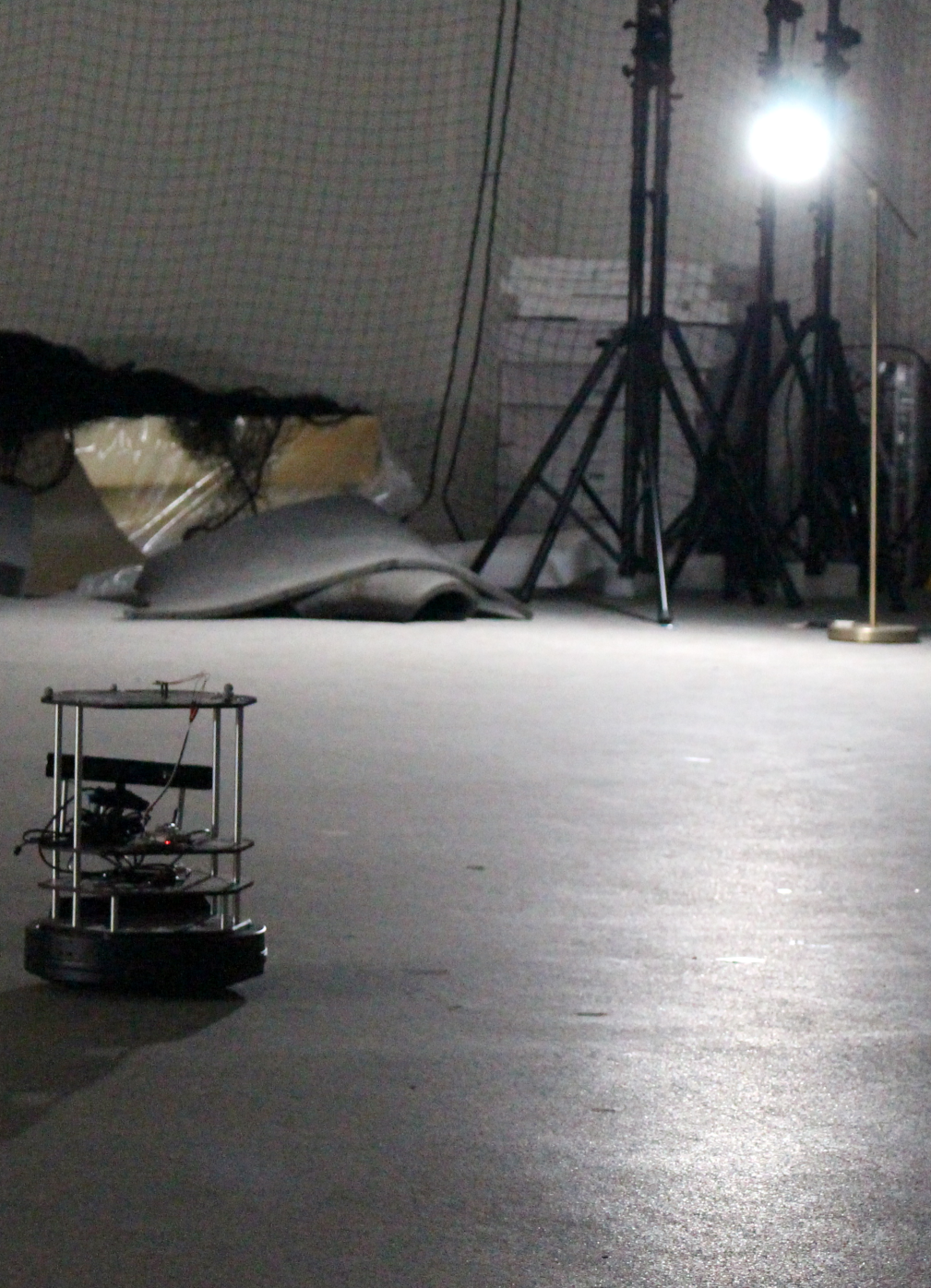}
    \end{subfigure}
    \caption{Left: {\bf Overall workflow.} {\sc Environment} is the robot's sensing space (\eg a lake) and {\sc Environment model} is its current understanding of the environment (modeled here by a Gaussian process). We introduce a learned {\sc Parameter Selection} agent that selects parameters (\eg number of rollouts) of {\sc Action Selection}, which determines the actions the robot should take (a POMDP solver in our work). The process repeats until a fixed number of informative path planning iterations are completed. Right: {\bf Field Trial.} We demonstrate that our approach transfers to a previously unseen real robot setup using a light sensor mounted on a mobile robot}\label{fig:hero}
    \vspace{-0.2in}
\end{figure}

A typical formulation for the IPP problem is a Partially Observable Markov Decision Process (POMDP) whose (exact) solution is the (optimal) action for each possible belief over the world states ({\sc Action Selection} in \cref{fig:hero}). As the robot executes these actions in its surroundings (the {\sc Environment}), it samples the phenomenon of interest and builds a representation ({\sc Environment Model}) which becomes an input to the {\sc Action Selection} process.

\noindent \textbf{Contributions} We introduce a reinforcement learning-based \textsc{Parameter Selection} agent that incorporates knowledge of the environment and sampling objective to learn and automatically set the parameters of the policy generation algorithm for each IPP iteration. We collect a corpus of 1080 publicly available highly diverse field robot trajectories some of which we use for training and the rest for evaluation.

We show that learning-based parameter selection results in an $9.53\%$ mean improvement in the cumulative reward across varied environments when compared to to end-to-end learning-based approaches for action selection. We also find a $3.82\%$ mean improvement in the cumulative reward when compared to a baseline, while expertly chosen parameters only achieve a $0.3\%$ increase in mean cumulative reward when compared to the same baseline. A field experiment illustrates the superior performance of our system out of the box.

\section{Background}

We frame the IPP problem as a POMDP and the problem of parameter selection as a Markov Decision Process (MDP) which we solve by training a policy network through reinforcement learning.
We discuss related work in IPP and the associated objectives to be maximized through information gathering.
Following this, we briefly discuss reinforcement learning as a solution for complex MDPs and how reinforcement learning and robotic information gathering have been previously combined and compare to our proposed system.

\subsection{Informative Path Planning}
Planning for robotic information gathering has been extensively studied in literature and tested in field systems.
Recently, rollout-based solutions have gained popularity to perform non-myopic planning for challenging missions including multi-robot deployments~\cite{seth_jfr, best_dec-mcts_2019}, multiple objectives~\cite{liu_paper}. 
Field tests of these systems show great potential for widespread uses~\cite{fossum_information-driven_2018,seth_jfr,joao_iser, das_sampling}.
Many planners (including this work) commonly use a Gaussian process (GP) to model the environment. 
The main benefit of GPs is that they allow for a non-parametric representation of the environment and contain uncertainty quantification of a predicted value~\cite{seth_jfr,denniston_icra_2021,borenstein_bayesian_2014,Marchanta}. 

Bayesian optimization is a popular method for finding the maxima of a black box function. IPP and Bayesian optimization share a common theoretical framework, especially when the maximal values of a spatial field are desired~\cite{Marchanta,borenstein_bayesian_2014}. 
In this work, we test the performance of learning-based parameter selection for IPP with the expected improvement and probability of improvement Bayesian optimization objective function.

Rollout-based algorithms are popular for IPP as they allow for anytime computation of plans on constrained hardware~\cite{seth_jfr,best_dec-mcts_2019}.
In particular, Monte Carlo tree search-based solutions allow for long horizon plans to be built without requiring exhaustive search at all branch levels~\cite{Marchanta}.
Formulating IPP as a POMDP and solving using various rollout-based solvers has been shown to be effective~\cite{Marchanta,choudhury_adaptive_2020}. We adopt a rollout-based approach in this work.

\subsection{Reinforcement Learning}

Policy gradient methods are a class of reinforcement learning methods where the robot policy is represented as a distribution over actions. 
Proximal Policy Optimization (PPO) is a popular policy gradient method used to solve robotics tasks, and has shown success in many continuous control tasks~\cite{schulman2017PPO}.

In this work, we compare against two end-to-end RL systems to demonstrate the need for combining classical IPP planning techniques with a learning system.
Long short-term memory (LSTM)~\cite{hochreiter1997lstm,sundermeyer2012lstm} is a neural network architecture that has feedback connections and a hidden state, which enables it to process sequential data.
Graph Neural Networks (GNNs)~\cite{monti_geometric_2016,GNNs2008} are a family of neural networks that can process data represented by graphical data structures. 
These networks can readily be applied to IPP because the LSTM cells can capture the temporal relationship whereas the GNNs can capture the spatial relation between the data samples.

\subsection{Combining Informative Path Planning and Reinforcement Learning}
Planning and optimization have been combined with RL for robotics applications~\cite{yamada2020mopa, xia2020relmogen}.
Specifically, informative path planning has been augmented with RL in different ways. 
In previous work, a learned system has been used to select between different adaptive sampling policies~\cite{choi_adaptive_2021}.
Recurrent policies have also been deployed directly to IPP -  the policy is trained end to end such that the policy network selects the best action at each step on a graph~\cite{wei_informative_2020}.
RL on graphs has shown great promise in information gathering tasks.
Particularly, work has been done to train agents to automatically predict exploratory actions by encoding a SLAM pose graph as the input to a graph neural network~\cite{chen_autonomous_2020}. RL has also been combined with rollout-based solvers for IPP by learning to improve the policy used for tree exploration and rollouts by learning to predict the value function~\cite{ruckin_adaptive_2021}.

\section{Formulation and Approach}
IPP can be formulated by maximizing some objective function $f$ over a set of measured locations $\sensedlocations$ and corresponding set of measured values $\sensedvalues$ subject to some cost $c(\sensedlocations)$ less than a fixed budget $B$, which defines the number of IPP iterations. 
Typically, the trajectories the robot follows to gather data are created in real-time by interleaving planning and action during   operation.
Formally, IPP can be described as finding the optimal set of sensing locations $\sensedlocations^* = \argmax_{\sensedlocations \in \Phi} f(\sensedlocations) | c(\sensedlocations) \leq B$ where $\Phi$ is the space of robot trajectories~\cite{Hollinger2014,rayas_icra_2022}. 

We separate the planning space from the sensing space.
The set of all robot poses that can be visited is $\gtlocations$ while the set of all locations that can be sensed is $\gtsensedlocations$. 
At each $\sensedlocation \in \sensedlocations$ the robot can sense a value $\sensedvalue \in \sensedvalues$. 
A sensing function $\sensingfunction: \gtlocations \times \gtlocations \rightarrow \gtsensedlocations$ defines the set of locations that can be sensed by moving from one robot pose to another.
This is to allow sensors that sense multiple points from a single location, such as a camera, or allow the robot to capture data while transitioning from one location to another~\cite{rayas_icra_2022}.

We approach the problem of finding an optimal path in two steps, namely planner parameter selection and action planning. 
At each IPP iteration the \textsc{Parameter Selection} agent chooses parameters to configure the planner which then solves the underlying IPP POMDP. The \textsc{Parameter Selection} agent is trained using RL, to allow it to perform long horizon planning over the course of a trajectory.

We model the environment using a Gaussian process (GP).
Gaussian processes use a kernel function to determine the effect of the value of neighboring points on the predicted value at a specific point~\cite{Rasmussen2006}. A GP can be used to estimate the mean $\mu(\sensedlocation)$ and variance $\sigma^2(\sensedlocation)$ at a specific location $\sensedlocation$. We adopt the squared exponential kernel which is standard for IPP~\cite{seth_jfr,Marchanta}.

\subsection{Objective Functions}\label{sec:objective_functions}
We compare three classes of objective functions which are used for IPP to demonstrate the performance of our \textsc{Parameter Selection} agent.
Entropy-based coverage objective functions are used when a robot has to spatially cover an area and maximally reduce the uncertainty about the concentration in the model.
Bayesian optimization objective functions are used when the maximal values of a spatial field are desired, or it is desirable to spend more time exploring higher concentration areas.
These objective functions take a sensed location, $\sensedlocation_i$, and produce a scalar reward, $r_i = f(\sensedlocation)$, determined by the utility of sensing a value at that location.
Given a set of points produced by the sensing function $\sensedlocations_j = \sensingfunction(\location_{j-1},\location_{j})$,  the objective function is denoted by $f(\sensedlocations_j) = \Sigma_{\sensedlocation_i \in \sensedlocations_j} f(\sensedlocation_i)$.

\noindent \textbf{Entropy-Based Coverage} A common objective function is {\bf entropy}, which relates to how well the model spatially covers the workspace \cite{guestrin_near-optimal_2005,low_information-theoretic_2013}. 
\begin{equation}\label{eq:entropy}
   \objectivefunction_\t{en}(\sensedlocation_i) = \frac{1}{2} log(2 \pi e \sigma^2(\sensedlocation_i))
\end{equation}

\noindent \textbf{Bayesian Optimization} 
Bayesian optimization objective functions seek to find the maxima of a concentration. 
Two common objective functions are expected improvement and probability of improvement~\cite{Marchanta,probability_of_improvement}.
Both {\bf probabilty of improvement} and {\bf expected improvement} use the notion of improvement, which is defined as $ I = (\mu(\sensedlocation_i) - max(\mu(\gtsensedlocations)))$. This improvement is used when calculating a Z-score defined as $Z =\frac{I}{\sigma^2(\sensedlocation_i)}$. 

{\bf Probability of improvement} is an objective function which seeks to find areas which have a high probabilty of having a measured value higher than the largest previously measured value. 
Probabilty of improvement is defined as 

\begin{equation}\label{eq:pi}
    \objectivefunction_\t{pi}(\sensedlocation_i) =  
      \Phi (Z) 
\end{equation}
where $\Phi$ is the CDF of the normal distribution. 

{\bf Expected improvement} differs from probability of improvement by incorporating how much the new measured value is expected to improve over the previously seen maximal value, rather than just looking at the likelihood of it improving.
It has been used in IPP as an objective function when the maxima of a spatial field are desired~\cite{rayas_icra_2022,MarchantEI}.

\begin{equation}\label{eq:ei}
      \objectivefunction_\t{ei}(\sensedlocation_i) =
      I\Phi(Z) + \sigma(\sensedlocation_i)\phi(Z)
\end{equation}
where $\Phi$, $\phi$ are the CDF and PDF of the normal distribution.

\subsection{POMDP Solver}
In formulating IPP as a POMDP, the state of the environment (the ground truth $GT$) is the hidden state and sensor measurements $\sensedvalue_t = GT(\sensedlocation_t)$ are the observations~\cite{borenstein_bayesian_2014}. 
The actions are feasible poses that can be reached from the current robot pose $\location_t$.
The belief state is defined by the Gaussian process over the observed values and locations, $GP(\sensedlocations_{1:t},\sensedvalues_{1:t}$), and is augmented with the known robot pose $\location_t$.
The reward is defined by the objective function $f(x_t)$, with future rewards discounted by $\gamma^d$.
To solve this POMDP we use a rollout-based solver POMCP~\cite{silver_monte-carlo_2010} - a Monte Carlo tree search-based algorithm that continually improves the estimate of the value of each action by using rollouts. 
The rollouts simulate the robot taking actions and receiving observations to a fixed horizon $h$ and update the value function~\cite{silver_monte-carlo_2010}.
We adopt the t-test heuristic to allow the robot to confidently take multiple steps from a single POMCP plan~\cite{denniston_icra_2021}. 
Because the observations $GT(\sensedlocation)$ for unseen locations are not known during planning, the predicted value from the GP conditioned on previous observations is used~\cite{Marchanta}.

\begin{table}[h!]
\centering
\begin{tabular}{ m{5em} m{6em} m{15em} }
\textbf{Parameter} & \textbf{Range} &\textbf{Meaning} \\
\hline \hline
POMDP Rollouts & $[10,300]$ & Amount of rollouts to use from the POMDP root node\\ 
\hline
POMDP Solver $\gamma$ & $[10^{-1},.99]$  & Planner Discount for future rewards \\  
\hline
T-Test Value & $[10^{-3},0.4]$ & Determines when the planner can take multiple steps from a single plan\cite{denniston_icra_2021} \\
\hline
Max Planning Depth $h$ & $[3,15]$ & The number of generator calls from the root node before a rollout is terminated
\end{tabular}
\caption{{\bf Parameters for the POMDP solver which are chosen by the parameter selection agent.} These parameters critically impact the effectiveness of the POMDP solver and are difficult to set in the field for non-technical users.}\label{tbl:parameters}
\vspace{-0.22in}
\end{table}

\subsection{\textsc{Parameter Selection} Agent}
Rollout-based POMDP solvers are effective in computing long horizon non-myopic plans for IPP~\cite{Marchanta}.
They have a suite of parameters that can be modified to greatly improve the reward gained by the planner thereby drastically impacting its performance~\cite{denniston_icra_2021}. 
These parameters can be set statically but do not correspond to physical quantities so they are difficult to set even by experts. Previous work has shown that these parameters can be more effectively set if they are changed at each iteration of the planner~\cite{denniston_icra_2021}.
We focus on the parameters in \cref{tbl:parameters} because of their difficulty to set by a human operator and importance to the quality of the solution.

We define the problem of setting the parameters for a POMDP solver as an MDP with a stochastic reward obtained by executing the POMDP solver's policy in the environment.
The \textsc{Parameter Selection} agent's actions at time step $t$ are the parameters $\theta$ sent to the POMDP solver, and can be represented as $\theta = a_t \sim \pi_{\phi}(\cdot |s_t)$ with state (robot measurement and locations) $s_t \in \mathcal{S}$ and action (POMDP solver parameters) $a_t \in \mathcal{A}$, where $\phi$ denotes the parameters of the policy $\pi$ (policy network weights). 
The POMDP solver uses parameters $\theta$ to plan and execute an action for the robot. 
On execution in the training environment, the \textsc{Parameter Selection} agent receives a reward $r_t=R(s_t,a_t)=f(\sensedlocation_t)$ from the environment.
To solve this MDP, we employ proximal policy optimization (PPO), a widely adopted policy-gradient RL algorithm, which trains a network to directly output the selected parameters from the given state~\cite{schulman2017PPO}.
PPO trains a policy network which can be used later for task inference.

A problem that arises in using RL for IPP is the lack of a clear fixed length vector input for the policy network as the agent needs to consider a variable length history of samples.
Typically, in IPP the agent uses information from all gathered samples to determine the value of the next action.
In the POMDP solver, this is is done by using a GP which allows the solver to reason about the sensed value at a location and the uncertainty associated with the sensed value at that location. We propose two approaches for the policy network. 

\noindent \textbf{LSTM} One solution is to use a recurrent neural network, such as a long short-term memory (LSTM), as the policy network. 
This architecture allows information from previous samples to be propagated forward to the current time step using the internal recurrent state. However, this method has a drawback for IPP as its structure does not capture the spatial relationship between points naturally, but instead captures a sequential relationship. We employ a  network consisting of two 256 neuron layers followed by a 256 neuron LSTM layer.
The LSTM network only receives the most recent sample and the metadata because it can preserve information about the previous samples in its recurrent hidden state.

\noindent \textbf {GNN} A natural approach to encode spatial information between collected samples is to use a graph neural network (GNN). 
GNNs use the spatial relationships between samples by constructing a nearest neighbor graph~\cite{chen_iterative_2020}.
This allows the policy network to use a non-fixed input size and to incorporate information about the spatial relationship of samples into its decision making process.
The GNN consists of two inputs: the graph side, and the metadata side.
The graph input is constructed by  adding edges from each node representing samples to the 6 nearest samples and adding the local Cartesian coordinates to the edge features.

The graph side contains three Gaussian mixture model layers with a kernel size of 3~\cite{monti_geometric_2016}.
This layer type was chosen because it captures the spatial relationship between samples as edge features.
The graph layers are followed by a global additive layer which adds up all of the node embeddings and is followed by a single 192 neuron dense layer.
The metadata side consist of a single 92 neuron dense layer.
These layers are concatenated together and processed by a 256 neuron dense output layer.

\noindent \textbf{Ablations} We perform two ablations of our method.
Both are provided access to the metadata (the remaining number of rollouts, remaining fraction of rollouts, remaining number of IPP iterations, remaining fraction of IPP iterations, and a one-hot encoded vector of the selected objective function) of the current episode. One is additionally provided the last 10 samples as a fixed-length history.

We train each model on all objective functions simultaneously.
To train the \textsc{Parameter Selection} agent, we collect 50 environment steps on 32 parallel workers time training environment for each agent update step. 
We update the agent 250 times, which allowed convergence for all agents.
For all other parameters, such as learning rate, optimizer, etc, we use the default values for the implementation of PPO from Ray 1.6~\cite{ray}, as training appeared to be robust using these values.
For performance on diverse objective functions and environments, we normalize the reward by the mean and variance ($\mu_{obj}, \sigma_{obj}$) of the rewards for each objective function.
We also clip this reward to $[-3,3]$ and add a small survival bonus ($b_{survival}$ = 1). 
To discourage the agent from superfluously using generator calls, we multiply the number of generator calls used by a small penalty ($p_{gen} = 10^{-5}$) designed to encourage it to try to use as few calls as possible.
The resulting reward function is 

\begin{equation}
    R(r_t) = clip\left(\frac{(r_t-\mu_{obj}) }{\sigma_{obj}},-3,3\right) + b_{survival} - p_{gen} GC(t)
\end{equation}
where $GC(t)$ is the number of generator calls used in the POMDP solver at time $t$ and $r_t$ is the environment reward $r_t = f(\sensedlocation_t)$.
The environment reward is defined by computing the objective function using the environment value for the sample, not the Gaussian process belief, $\mu(x_i)$, which is used during planning for the POMDP solver.

\subsection{Data Acquisition}\label{sec:noaadata}
Learning-based approaches for field robotics have suffered from a lack of training data for the phenomena being sampled by the robot. 
We gather a large corpus (1080 trajectories) of publicly available data from ocean surveys performed by ocean glider robots tasked with measuring conductivity, salinity, and temperature. 
We use this dataset to train the \textsc{Parameter Selection} agent and show that by training on this corpus, our approach generalizes to other environments and phenomena (\eg chlorophyll sampling in small-scale lakes) on which the models are not trained.

We assimilated data from three publicly available sources: The Southeast Coastal Ocean Observing Regional Association (SECOORA)~\cite{secoora_website_2021},  the Central and Northern California Ocean Observing System (CeNCOOS)~\cite{cencoos_website_2020}, and the  Integrated Ocean Observing System's National Glider Data Assembly Center (NGDAC)~\cite{ngdac_website}. Each of these organizations collects data in a variety of locations (\eg the Pacific and Atlantic ocean, Gulf of Mexico, \etc). We collect and assemble this data using NOAA's ERDDAP system~\cite{errdap_website}.

We gather trajectories collected between January 1st 2005 and July 25th 2020 and manually visualize each trajectory, discarding those which were too small or discontinuous. This results in a corpus of 1080 trajectories which include the latitude, longitude, depth and time values as well as the sensed values at those locations. 

We train all the models used in this paper on this dataset of robot trajectories. 
During training and evaluation, we subsample each trajectory to be the same spatial scale.
We also randomly select a sensor type (one of conductivity, salinity, and temperature) and objective function (\cref{sec:objective_functions}) for each sub-trajectory.

\section{Experiments}
To validate our approach we perform experiments in three simulated environments and a field scenario.
Via the simulation experiments, we show that our approach can be trained on a large corpus of data from one environment and perform well on environments not used for training. 
In the field scenario, we show that our system is easy to deploy and achieves superior results with no in-field parameter tuning.
In all experiments, the models are trained from the dataset described in \cref{sec:noaadata}.
We do not retrain the models for each environment.

\newcommand{\twidth}{5.1em}
\newcommand{\ttwidth}{10.6em}

\begin{table*}[t!]
\centering
\footnotesize
\begin{tabular}{
m{10em} |  m{\twidth} m{\twidth} | m{\twidth} m{\twidth} |  m{\twidth} m{\twidth} | m{\twidth}  m{\twidth}}
 & \multicolumn{2}{c}{Classical} & \multicolumn{2}{c}{End-To-End RL} & \multicolumn{2}{c}{Ablation} & \multicolumn{2}{c}{Proposed} \\ \\
 \textbf{Method} &  Naive POMCP & Empirically Tuned POMCP \cite{denniston_icra_2021} & GNN RL  & LSTM RL~\cite{wei_informative_2020} & Metadata RL-POMCP & Fixed-Length RL-POMCP  & GNN\hspace{2em}RL-POMCP & LSTM\hspace{2em}RL-POMCP \\
\hline
\raggedright{\textbf{Learned Model Sample Input}} & N/A & N/A & k-NN Graph & Previous Sample & None & Last 10 Samples & k-NN Graph & Previous Sample \\
\hline
\raggedright{\textbf{Uses POMDP Solver}}           & True & True & False & False & True & True & True & True \\
\hline
\raggedright{\textbf{Learned Model Chooses Environment Actions}}          & False & False & True & True & False & False & False & False \\
\hline
\raggedright{\textbf{Trained on Glider Data}} & False & False & True & True & True & True & True & True \\
\hline
\raggedright{\textbf{Parameter Selection has  Metadata}} & False & True & False & False & True & True & True & True \\

\end{tabular}
\caption{\textbf{Description of Compared Methods}. 
Naive POMCP is an POMDP solver using manually chosen conservative parameters and the t-test heuristic~\cite{denniston_icra_2021}.
Empirically-tuned POMCP improves upon Naive POMCP with domain-specific features and tuned parameters~\cite{denniston_icra_2021}.
Metadata RL-POMCP has access to the current task metadata. 
Fixed-Length RL-POMCP has access to the metadata as well as the last ten samples (locations and values).
Metadata RL-POMCP and Fixed-Length RL-POMCP demonstrate the need for the parameter selection module to use the complete set of samples.
LSTM methods have access to the previous sample and uses an LSTM with a hidden state to access information from previous environment steps.
GNN methods use the all of the current samples in a graph neural network. }\label{tbl:methods}
\end{table*}

\subsection{Simulation Experiments}

In the simulated environments, we compare cumulative reward once the robot takes 50 environment steps.
We compare the three objective functions defined in \cref{sec:objective_functions}.
We perform the simulation experiments in three types of environments. 
In each environment, we simulate a continuous environment by interpolating the datasets.
We populate the GP with a small amount of random samples before sampling.

{\sc Ocean-CTD} is derived from the dataset used to train the models (\cref{sec:noaadata}), a large corpus of publicly available glider robot trajectories.
A subset of the data is held out for testing and used to validate the performance of the algorithms on data similar to what the models are trained on. We test on 30 different trajectories for each of the three objective functions.

{\sc Lake-Chlorophyll} uses is a dataset of lake-scale lake surveys from a small underwater robot. This environment is different from the {\sc Ocean-CTD} environment in two ways. 
First, the distance between robot states is reduced to $3m$ from $25m$. 
Second, it uses chlorophyll fluorescence as the sensed quantity which is not a sensor type the models are trained on.
We use five different datasets collected in a small reservoir with three seeds each. 
This data was collected during tests of other systems using an underwater robot equipped with a chlorophyll fluorescence sensor.
This data was collected between 2017 and 2018 using lawnmower surveys of the lake while oscillating between a fixed depth and the surface. This environment is used to validate that the models can generalize and be used in environments they are not explicitly trained for. 
In tests with the {\sc Ocean-CTD} and {\sc Lake-Chlorophyll} environments, the robot (a simulated underwater vehicle) is able to move in 6 directions: up, down, left, right, forward and back and collects samples while traveling between locations.

{\sc Bay-Camera} is an environment in which the robot is simulated as a fixed height and fixed yaw drone over hyper-spectral orthomosaics collected from previous drone flights over a small bay in a lake. The simulated robot chooses from four actions in a plane (forward, back, left and right) and has a hyper-spectral camera which, unlike a chlorophyll fluorescence or salinity sensor does not measure a single scalar at each location, instead capturing a hyper-spectral image composed of one value per pixel corresponding to each location in the viewing frustum of the camera. Additionally, in this environment the agent only makes measurements at discrete locations, unlike the previous two in which it continues to sample while traveling from one location to another.  
Finally, this environment is not defined by a regular cube, but by the bounds of the orothomosaic which presents difficult choices for planning. 
We use two different variants of the {\sc Bay-Camera} environment with three seeds each.
This data was collected by a drone on a field research mission for studying chlorophyll fluorescence distribution in a lake.

We compare our method variations against two different classes of methods with two methods each, shown in \cref{tbl:methods}.
We compare four different variations of our method, which we call RL-POMCP, on each task.
The first, Metadata RL-POMCP, only has access to the task metadata, that consists of the remaining number of rollouts, remaining fraction of rollouts, remaining number of IPP iterations, remaining fraction of IPP iterations, and a one-hot encoded vector of the selected objective function.
The second, Fixed-Length RL-POMCP, has access to the metadata and the previous ten samples (locations and values). 
The third, GNN RL-POMCP, uses all measured samples (as input to a GNN, to indicate spatial correlation) as well as the metadata. 
The fourth, LSTM RL-POMCP, uses the current sample and internal recurrent state, as well as the metadata.

Metadata RL-POMCP and Fixed-Length RL-POMCP represent ablated versions of our system and demonstrate the need for considering the collected samples in parameter selection.
Metadata RL-POMCP is a minimal encoding which would allow for a minimal adaptive behavior.
Fixed-Length RL-POMCP allows parameter selection to consider a finite history of samples, in contrast to LSTM RL-POMCP and GNN RL-POMCP which allow for consideration of all samples. 

We compare these variations to two methods which do not use learning to select the POMDP parameters (Classical methods), and two methods which do not use a POMDP solver but directly learn to map from states to actions using reinforcement learning (End-to-End RL methods). 
The first classical method uses a manually chosen set of parameters. 
The second classical method uses the empirically tuned planner in \cite{denniston_icra_2021}, tuned on data from the \textsc{Lake-Chlorophyll} environment.
The two end-to-end RL methods use the LSTM and GNN in the same way as their RL-POMCP counterparts.
However, they directly choose actions, as opposed to selecting parameters for a POMDP solver which chooses the actions.

\begin{figure}[t!]
    \centering
        \begin{subfigure}{\columnwidth}
                \centering
                \includegraphics[width=\textwidth]{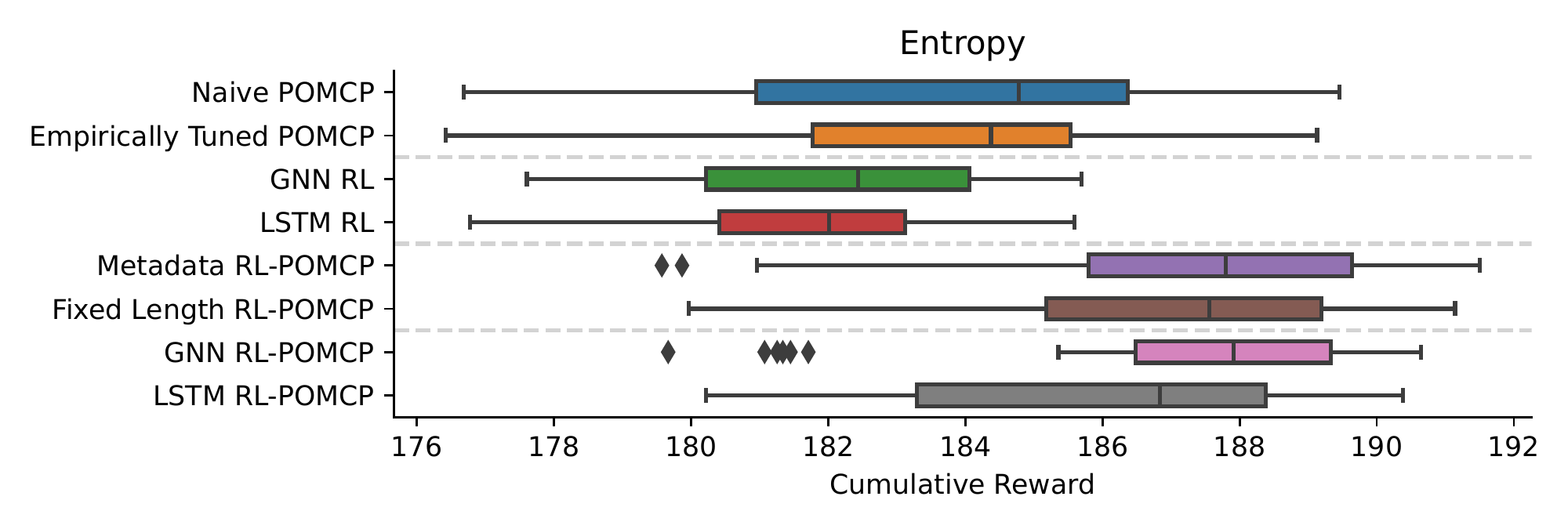}
        \end{subfigure}
        \begin{subfigure}{\columnwidth}
                \centering
                \includegraphics[width=\textwidth]{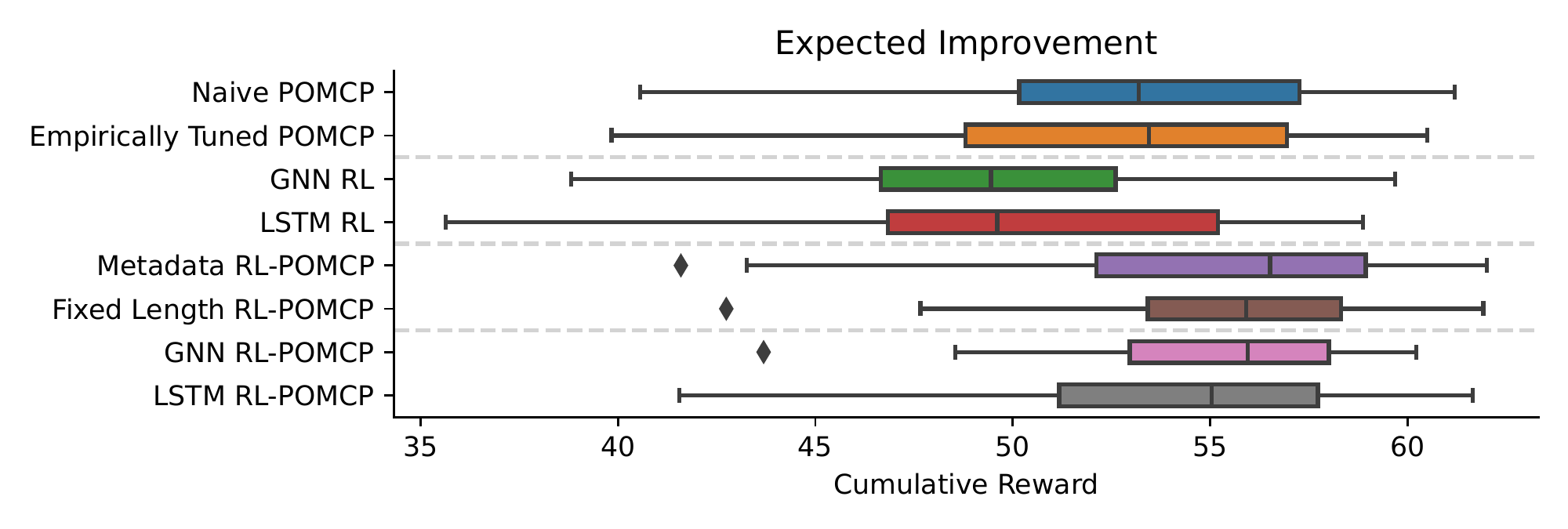}
        \end{subfigure}
        \begin{subfigure}{\columnwidth}
                \centering
                \includegraphics[width=\textwidth]{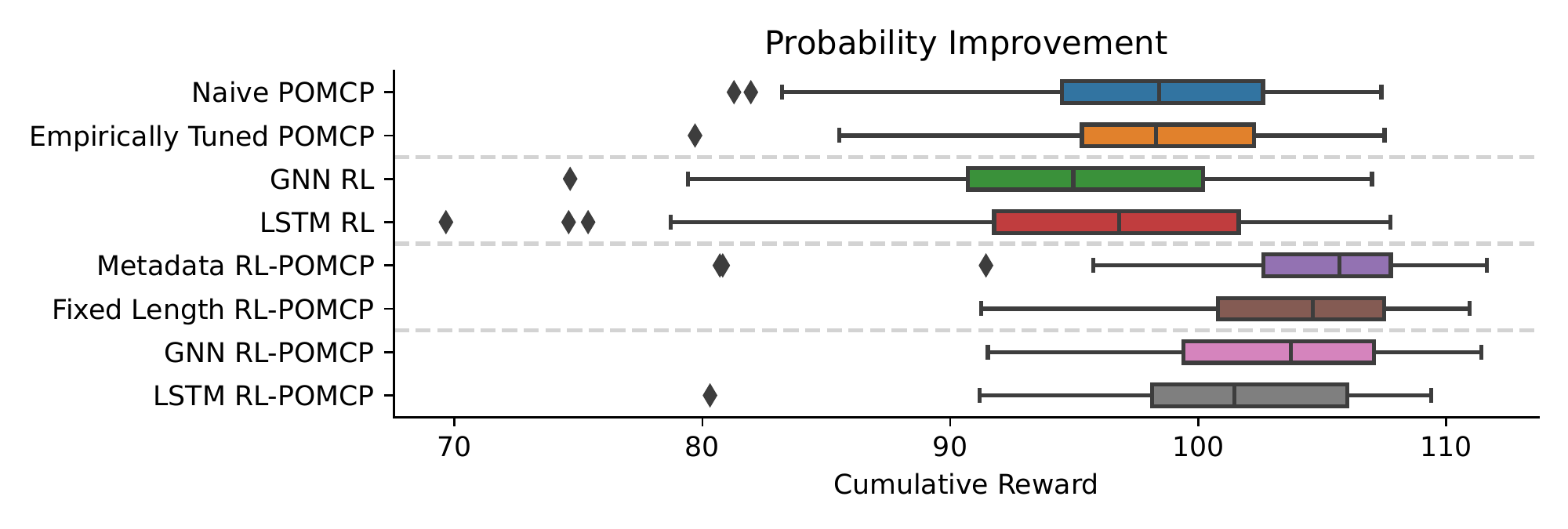}
        \end{subfigure}
        \caption{\textbf{Results on {\sc Ocean-CTD } from the data set of publicly available glider trajectories which the models were trained on.} We compare over 30 different trajectories and show the cumulative reward after 50 timesteps. We compare the three objective functions described in \cref{sec:objective_functions}. We find that our proposed methods outperform previous methods in all three objective functions.}\label{fig:noaa_results}
        \vspace{-0.2in}
\end{figure}

\begin{figure}[t!]
    \centering
        \begin{subfigure}{\columnwidth}
                \centering
                \includegraphics[width=\textwidth]{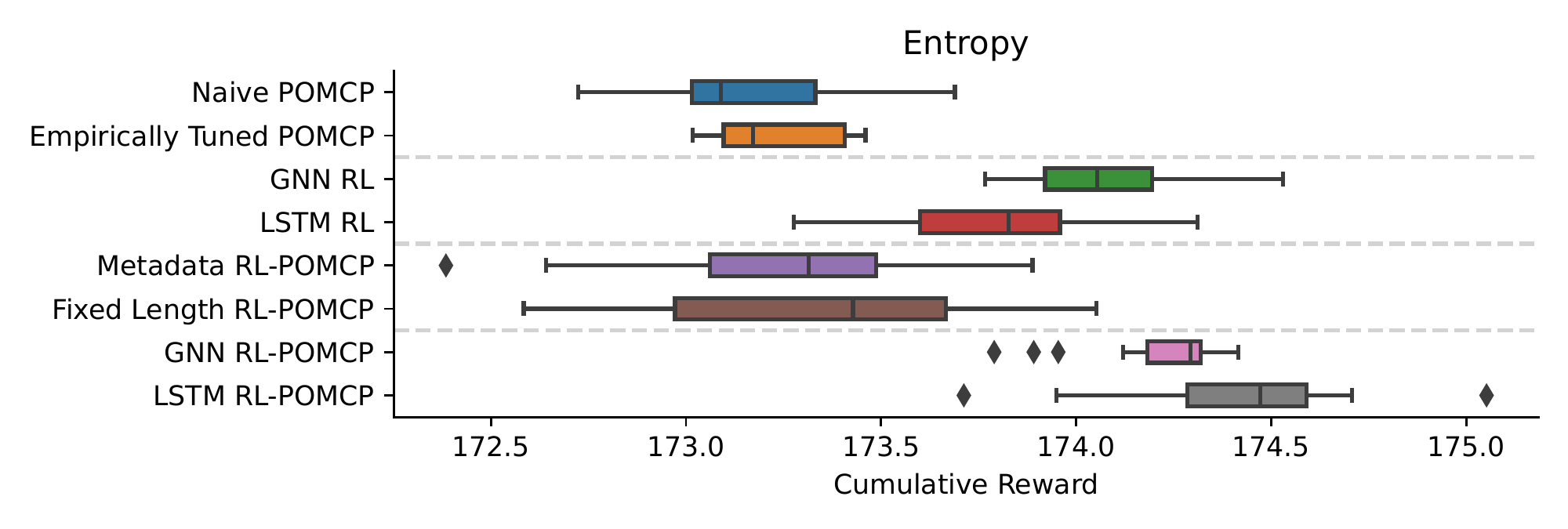}
        \end{subfigure}
        \begin{subfigure}{\columnwidth}
                \centering
                \includegraphics[width=\textwidth]{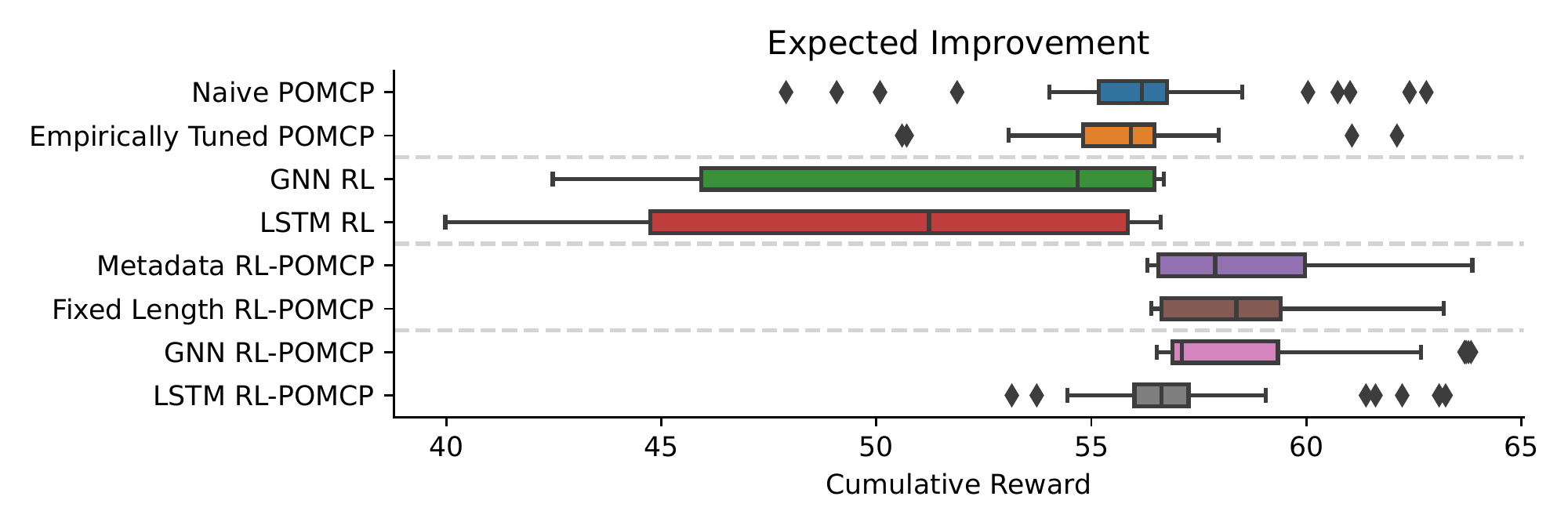}
        \end{subfigure}
        \begin{subfigure}{\columnwidth}
                \centering
                \includegraphics[width=\textwidth]{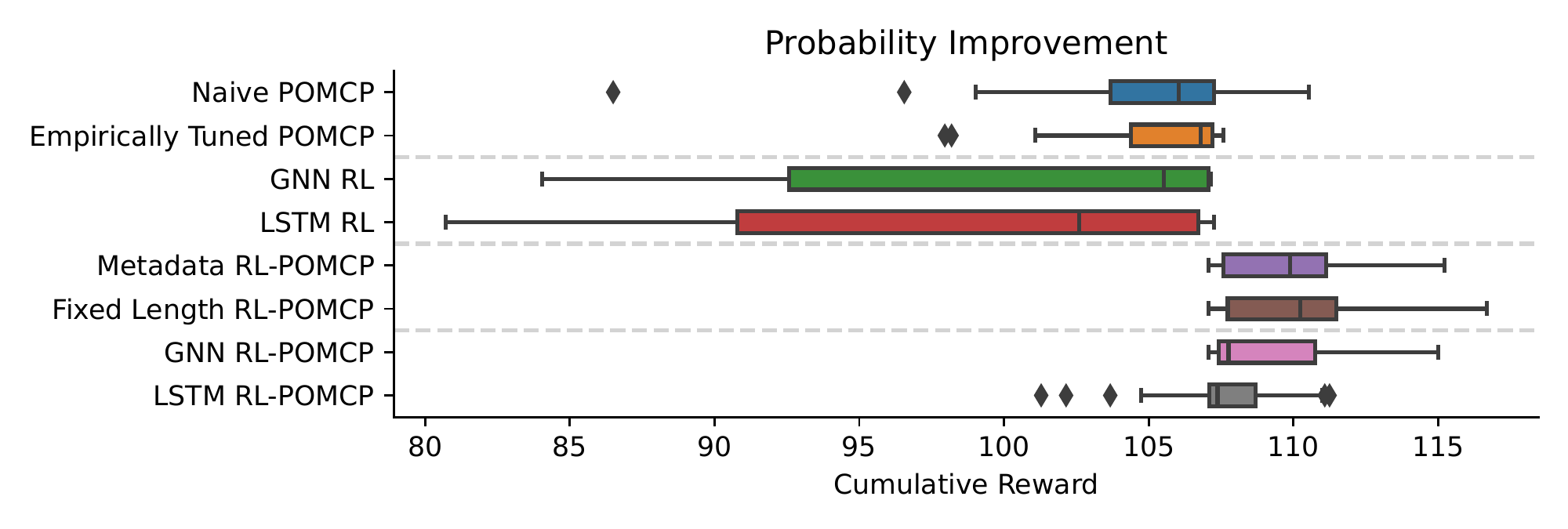}
        \end{subfigure}
        \caption{\textbf{Results on {\sc Lake-Chlorophyll} from a dataset of smaller lake trajectories using a chlorophyll fluorescence sensor that the models were not trained on.} We compare each objective function on 5 datasets with 3 seeds each. We find that GNN RL-POMCP outperform previous methods with the entropy objective function. All RL-POMCP agents perform well on expected improvement and probability of improvement}\label{fig:ecomapper_results}
        \vspace{-0.2in}
\end{figure}

\begin{figure}[t!]
    \centering
        \begin{subfigure}{\columnwidth}
                \centering
                \includegraphics[width=\textwidth]{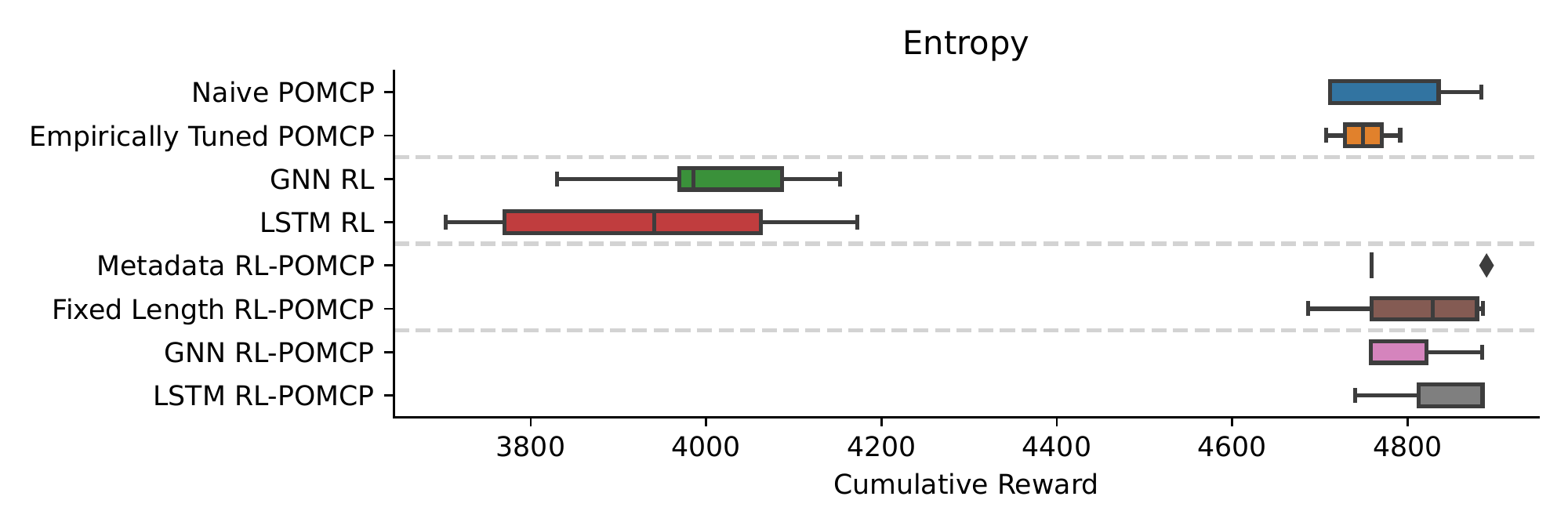}
        \end{subfigure}
        \begin{subfigure}{\columnwidth}
                \centering
                \includegraphics[width=\textwidth]{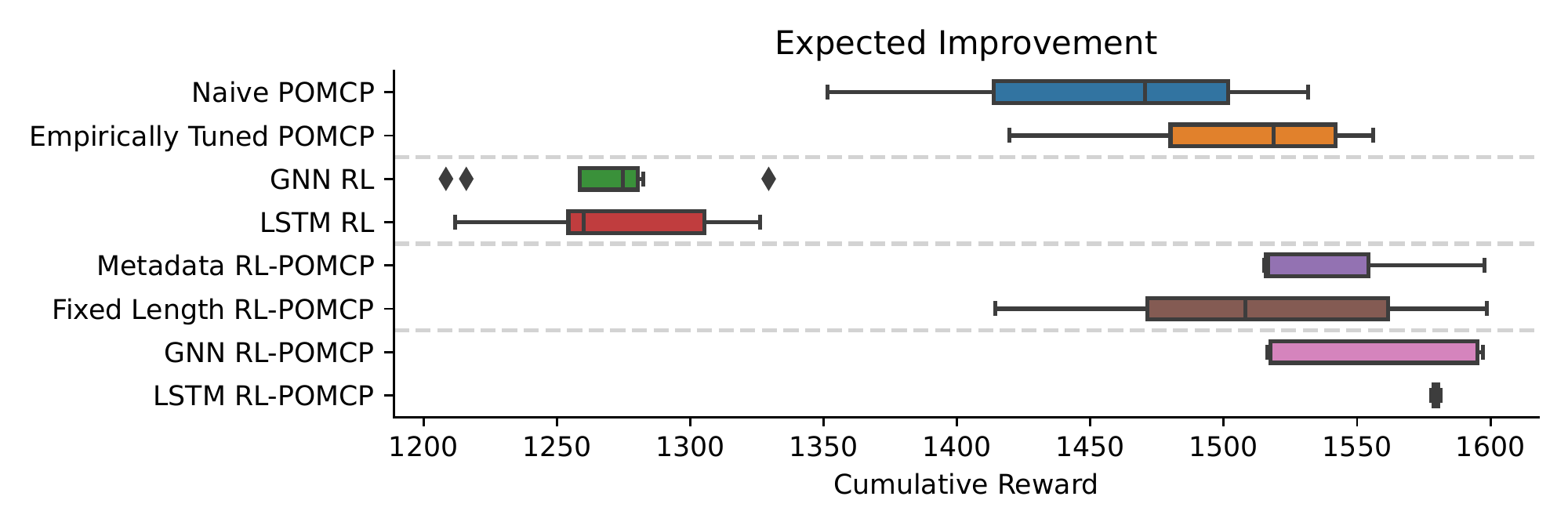}
        \end{subfigure}
        \begin{subfigure}{\columnwidth}
                \centering
                \includegraphics[width=\textwidth]{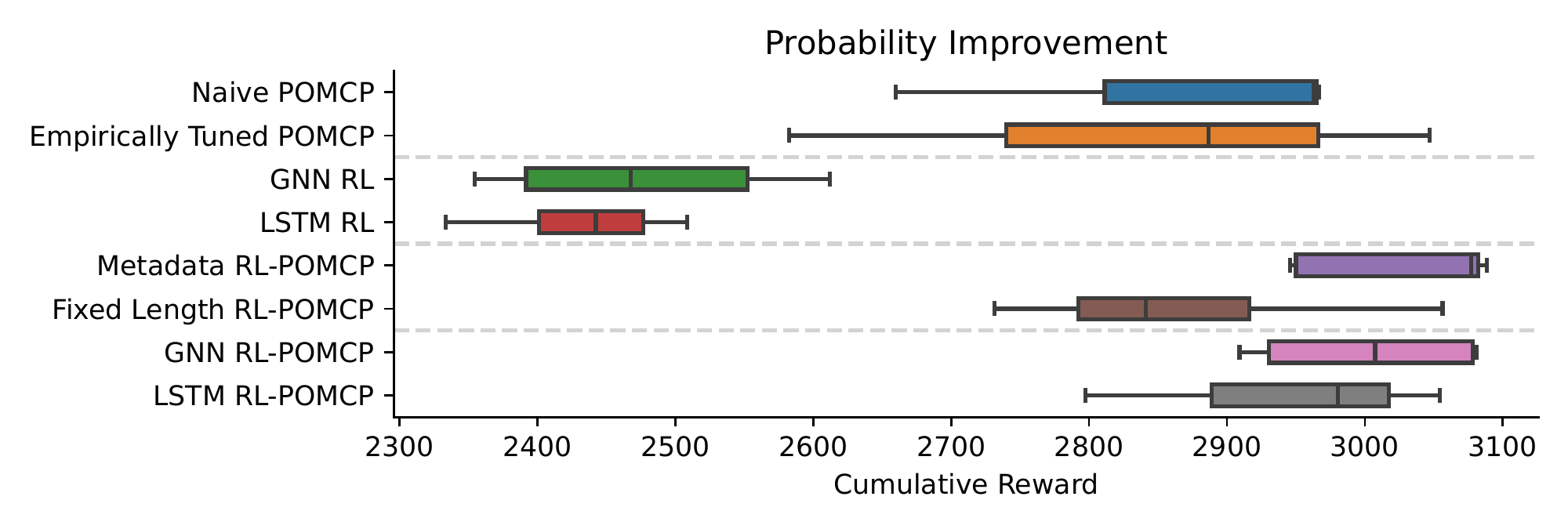}
        \end{subfigure}
        \caption{\textbf{Results for {\sc Bay-Camera}, a simulated drone flight environment.} In this environment, a drone is simulated flying over previously collected imagery. This imagery was collected using a hyper-spectral camera over a lake with algae mats on it. We evaluate on 2 images with 3 seeds each. Features which are not water were masked out. We find that our proposed methods perform comparably to prior methods on entropy. On the Bayesian optimization objective functions our methods perform well compared to the other approaches, demonstrating the ability of our approach to generalize to previously unseen sensor types. } \label{fig:drone_results}
\vspace{-0.2in}
\end{figure}

\subsection{Simulation Results}
\noindent \textsc{\textbf{Ocean-CTD}} The results for the \textsc{Ocean-CTD} environment can be seen in \cref{fig:noaa_results}. On the entropy objective function, the GNN, Fixed length, and Metadata RL models outperform the others.
For the expected improvement objective function we find that all the RL-POMCP models perform well. The naive POMCP performs as well as the POMCP model that was emprically tuned on the \textsc{Lake-Chlorophyll} dataset.
For the probability of improvement objective function, all RL-POMCP methods outperform the baselines. 
In all environments, the naive POMCP and empirically tuned POMCP methods are comparable, which is likely because the empirically tuned model is tuned on data from the \textsc{Lake-Chlorophyll} dataset. 

\noindent \textsc{\textbf{Lake-Chlorophyll}} The results for the \textsc{Lake-Chlorophyll} environment can be seen in \cref{fig:ecomapper_results}. 
This dataset is purposely chosen to show the generalization capabilities of our models when used on a dataset it was not directly trained on.
We find that our LSTM RL-POMCP model performs exceptionally well on the entropy objective function, followed by the GNN RL-POMCP model. 
In the expected improvement objective function, we find that both the LSTM RL-POMCP and GNN POMCP-RL models again performs well, and that the naive parameters and empirically tuned system performs well.
For the probability of improvement objective function, the metadata and fixed length RL-POMCP models perform the best, followed by the GNN and LSTM RL-POMCP models.

\noindent \textsc{\textbf{Bay-Camera}} The results for the {\sc Bay-Camera} environment can be seen in \cref{fig:drone_results}. 
This environment contains both a different sensor (a camera) and a different sensed quantity (hyper-spectral reflection on a lake) than the dataset the models were trained on.
We find that in Entropy and Expected Improvement the methods which just use a POMDP solver perform well and the methods which use end to end RL perform very poorly.
GNN RL-POMCP performs well on these two objectives but is outperformed by the empirically tuned system. 
For the probability of improvement objective function, all RL-POMCP methods perform well, except for the Fixed length RL-POMCP.
The naive POMCP performs slightly worse than the RL-POMCP methods.

\begin{figure}
    \centering
    \begin{subfigure}{.49\columnwidth}
       \includegraphics[width=\textwidth, trim={0.15in 0.15in 0.15in 0.15in},clip]{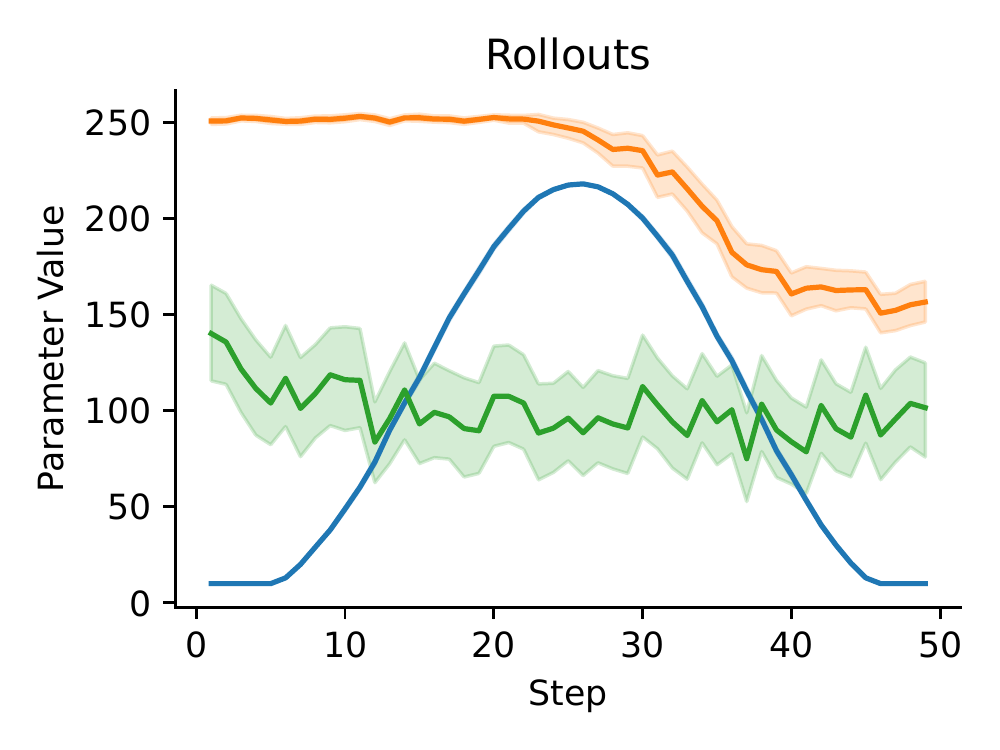} 
    \end{subfigure}
    \begin{subfigure}{.49\columnwidth}
       \includegraphics[width=\textwidth, trim={0.15in 0.15in 0.15in 0.15in},clip]{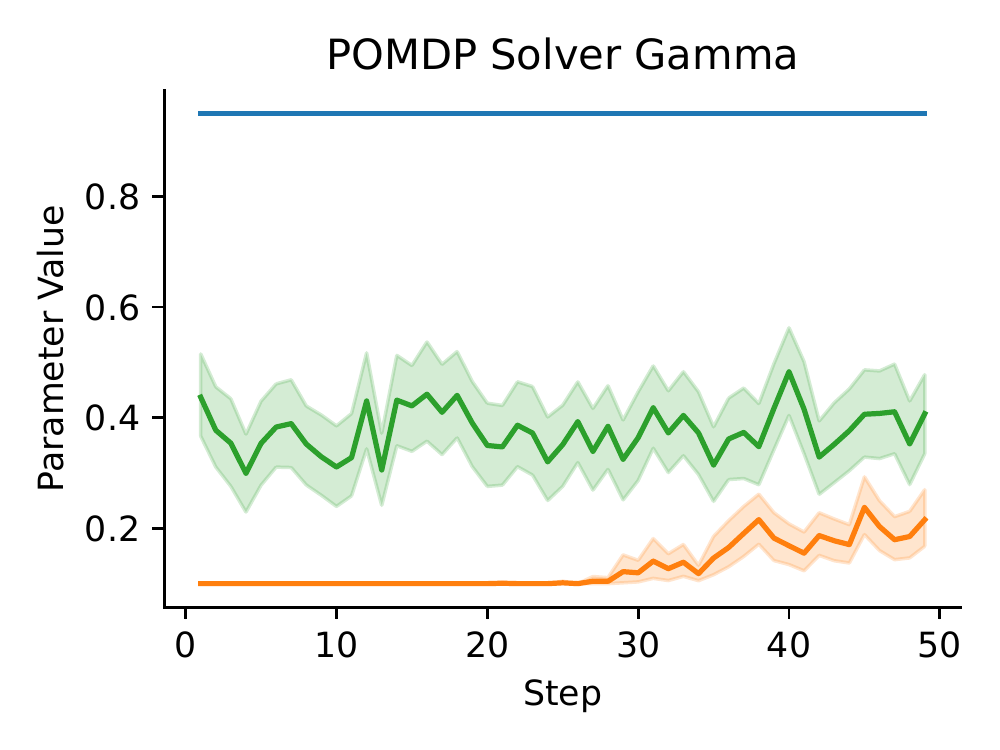} 
    \end{subfigure}
    \begin{subfigure}{.49\columnwidth}
       \includegraphics[width=\textwidth, trim={0.15in 0.15in 0.15in 0.15in},clip]{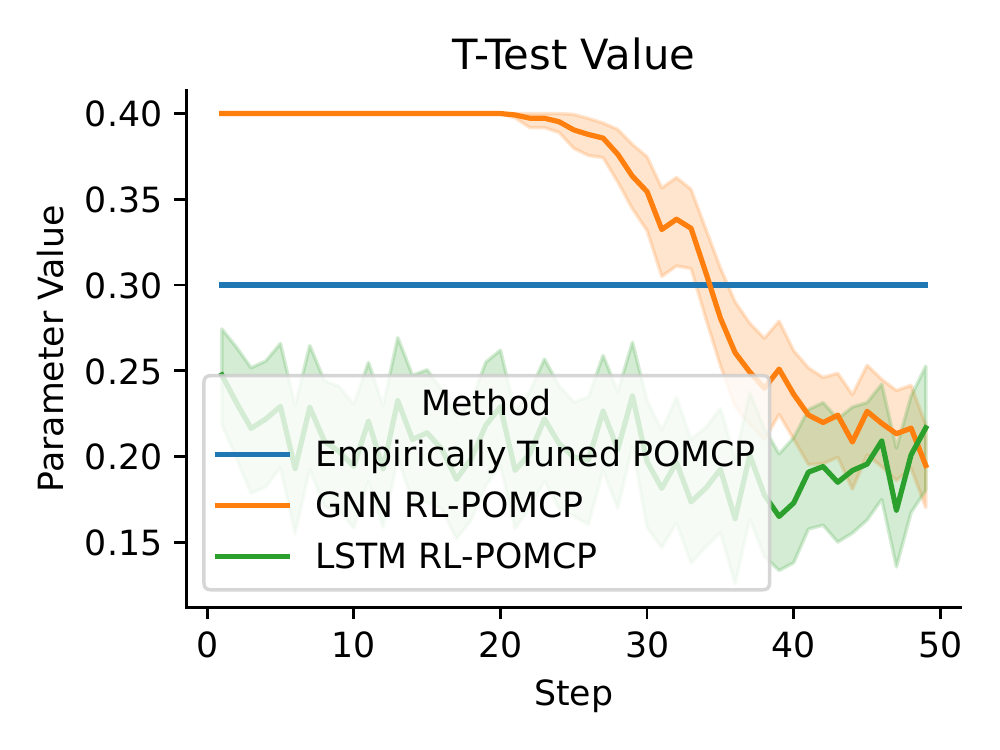} 
    \end{subfigure}
    \begin{subfigure}{.49\columnwidth}
       \includegraphics[width=\textwidth, trim={0.15in 0.15in 0.15in 0.15in},clip]{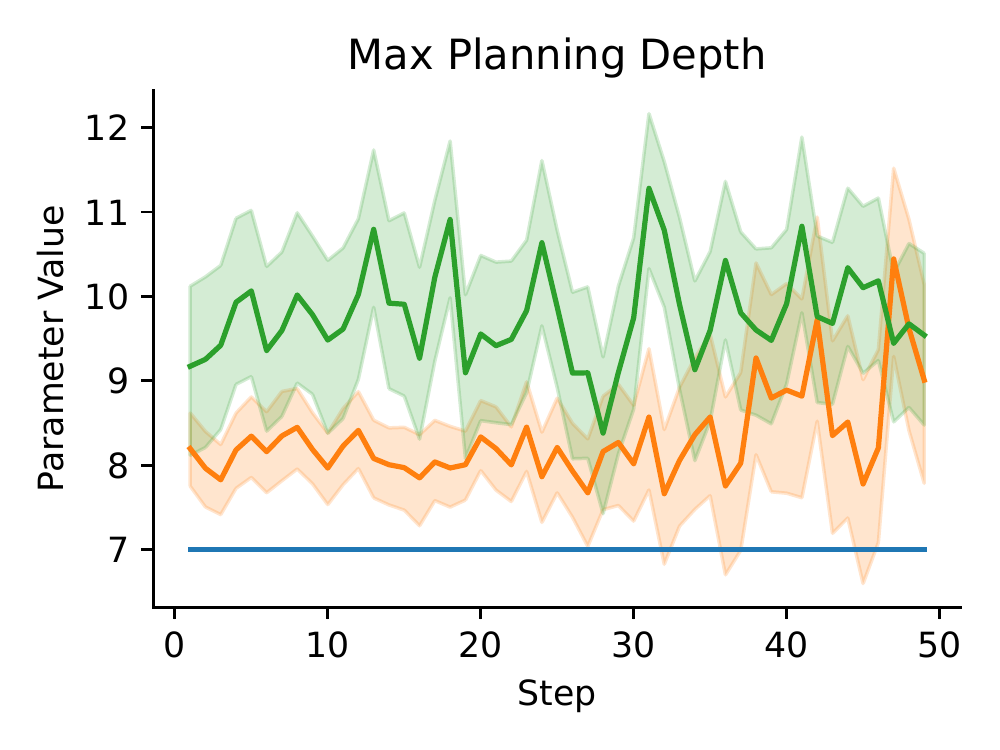} 
    \end{subfigure}
    \caption{Trajectories for each parameter across all evaluation environments for the expected improvement objective function (\cref{eq:ei}). LSTM RL-POMCP and GNN RL-POMCP learn distinct policies: GNN RL-POMCP many rollouts with a shorter horizon until the end when it tends to increase the horizon and reduce the number of rollouts, while LSTM RL-POMCP varies around a mean value the entire time.}
    \label{fig:param_trajs}
    \vspace{-0.1in}
\end{figure}

\noindent \textbf{Parameter Trajectories}
To show that the \textsc{Parameter Selection} agent learns a policy which varies the chosen parameters, we plot the parameter values over time in \cref{fig:param_trajs}.
This demonstrates that the agents do not simply chose a fixed set of parameters for all environment steps, but instead vary their parameters with the received measurements.

\begin{figure}[t!]
    \centering
        \begin{subfigure}{.49\columnwidth}
                \centering
                \includegraphics[width=\textwidth,trim=2.5cm 0.5cm 2.5cm 0.3cm,clip]{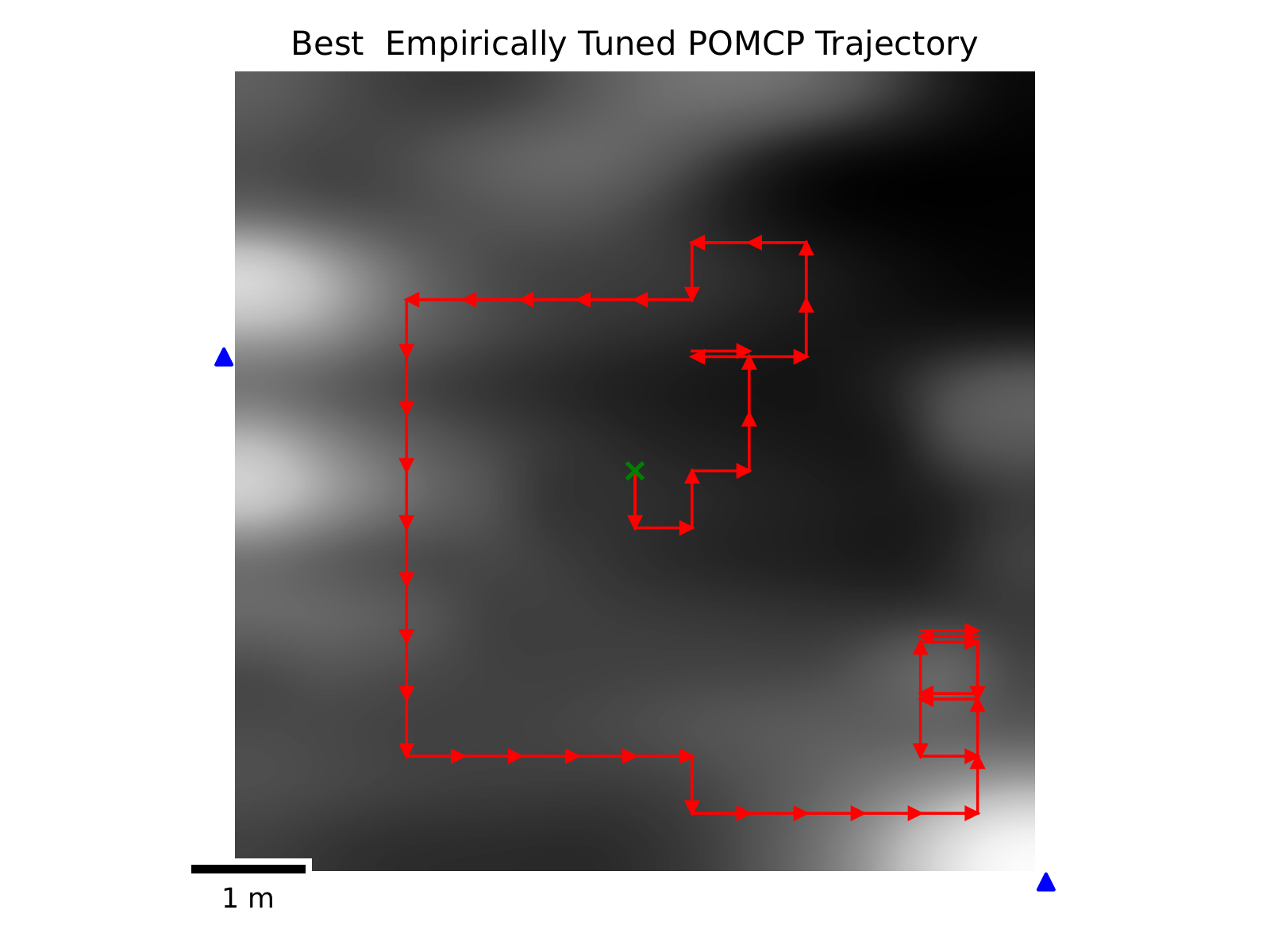}
        \end{subfigure}
        \begin{subfigure}{.49\columnwidth}
                \centering
                \includegraphics[width=\textwidth,trim=2.5cm 0.5cm 2.5cm 0.3cm,clip]{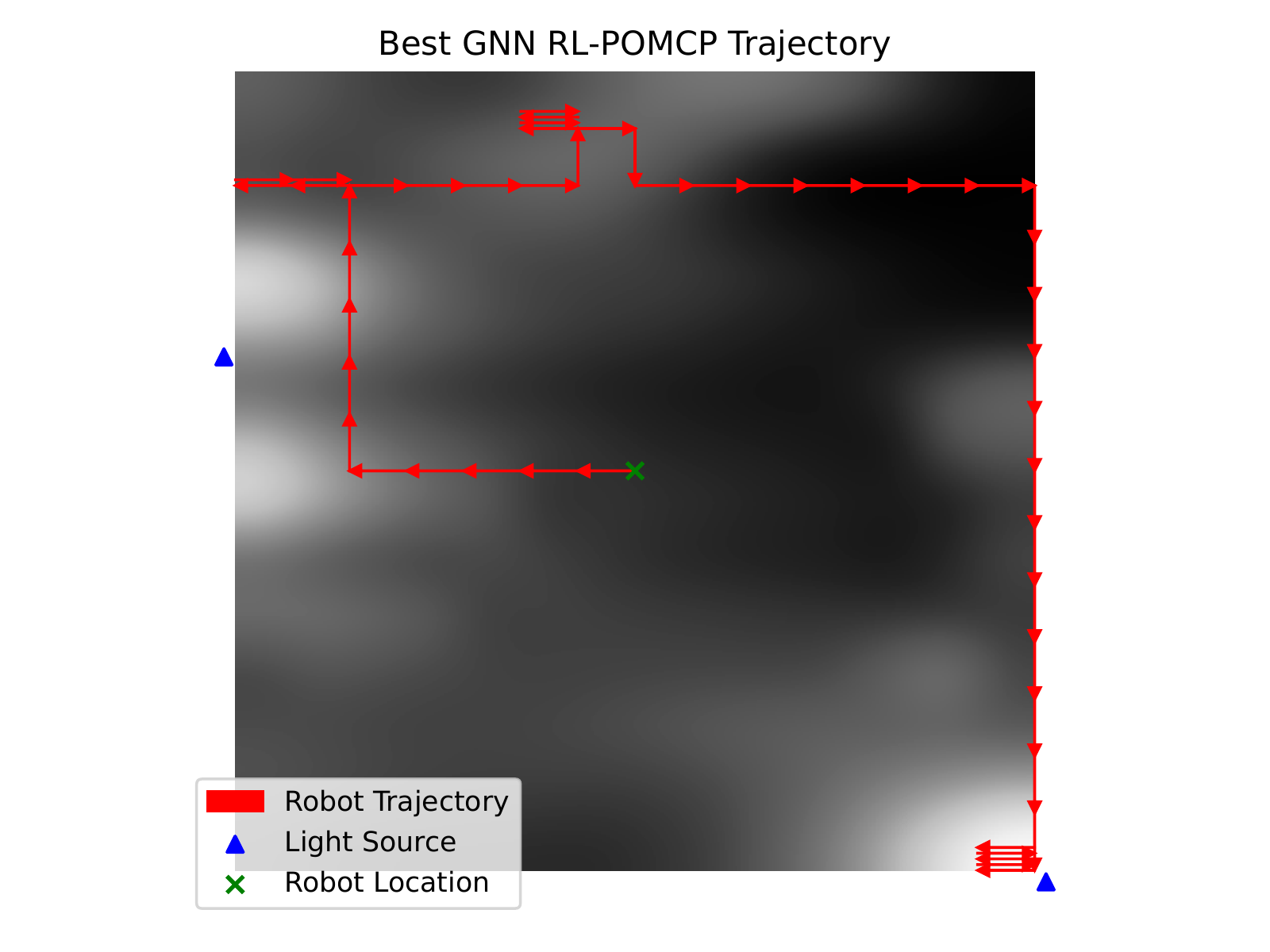}
        \end{subfigure}
        \begin{subfigure}{\columnwidth}
                \centering
                \includegraphics[width=\textwidth,trim=0.3cm .3cm .3cm .4cm,clip]{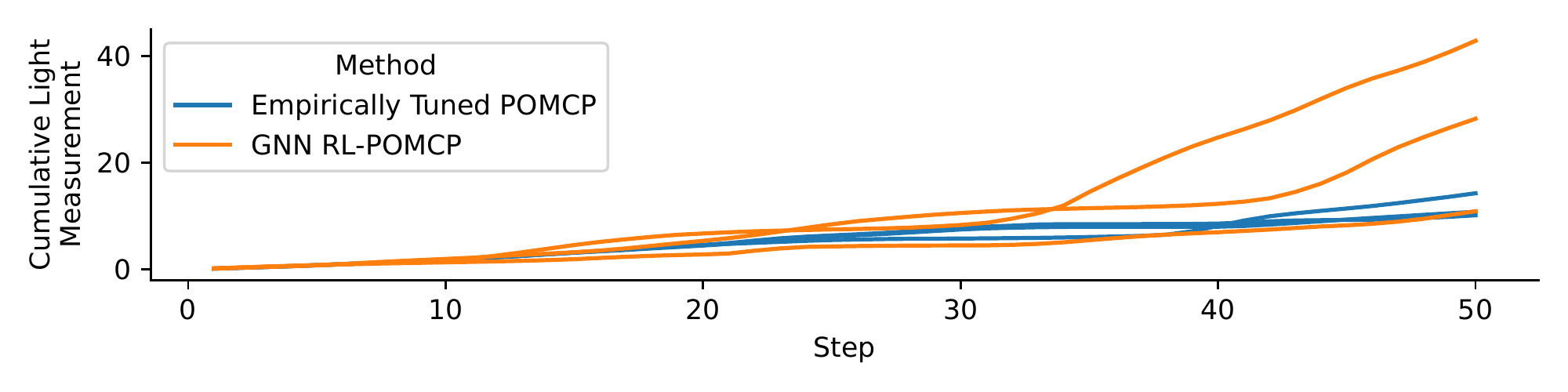}
        \end{subfigure}
        \begin{subfigure}{\columnwidth}
                \centering
                \includegraphics[width=\textwidth,trim=0.3cm .2cm 0 .5cm,clip]{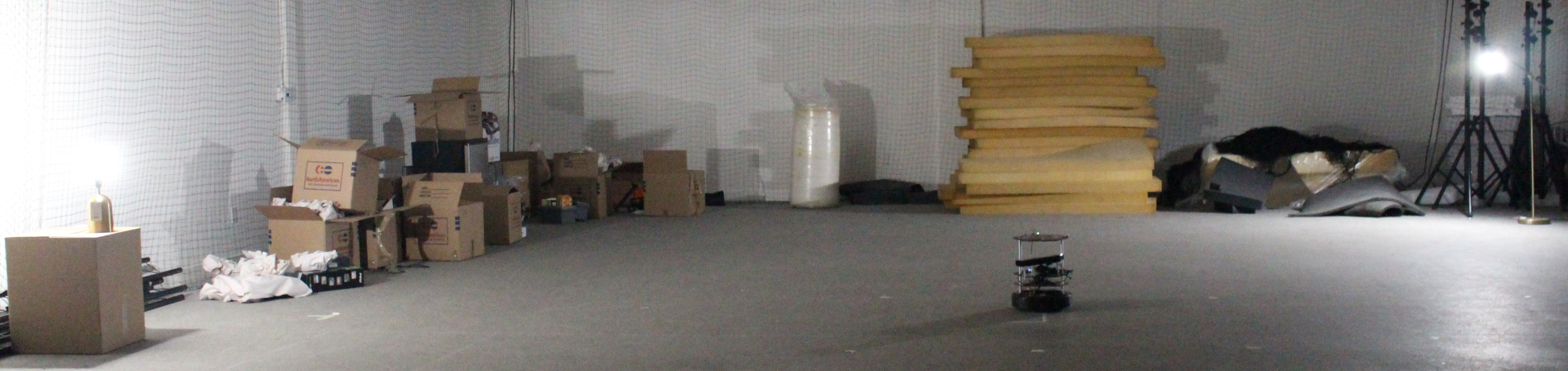}
        \end{subfigure}
        
        \caption{\textbf{Field Results.}
        Top: Comparison of trajectories for Empirically Tuned POMCP and GNN RL-POMCP deployed on a real robot taking light measurements. The red arrows show the trajectory taken by the robot. Two lights were deployed with existing light in the area.
        Middle: Comparison of the sum of the measurements taken over the number of actions taken by the robot and a picture of the experimental setup.
        Bottom: Experimental setup used with two placed lights.
        }\label{fig:field}
\vspace{-0.3in}
\end{figure}

\subsection{Field Experiment}

We deploy our system on a small mobile robot (Turtlebot2) with a light visible sensitive sensor. 
To model a complex concentration, we deploy two 1440 lumen light sources: one omnidirectional light source, and one directed one.
The robot moves in a 15x15 grid, spaced 0.5$m$ apart. The robot takes 50 steps in the environment (actions and measurements).
Based on simulation results, we deployed GNN RL-POMCP because of its superior performance in unknown environments with the expected improvement objective function.
The trajectory the robot took using both the GNN RL-POMCP and Empirically Tuned POMCP can be seen in \cref{fig:field}, as well as the cumulative sum of measurements over three trials for each method. 
Our system is able to find high lighting areas even in the presence of complex distributions and reflections against the walls.
Qualitatively, we find that the system delivers on our goal of easy deployment for IPP. In the field, we had to select the area for the robot to move in in, the number of measurements to take, and the objective function. 
This is comparable to the number and type of parameters in a lawnmower survey. All the parameters are comprehensible to a non-technical user, in contrast to the typical parameters in \cref{tbl:parameters}.

\section{Conclusion}
We presented a reinforcement learning-based approach to automatically tune the parameters governing action selection for a broad class of adaptive sampling techniques collectively called informative path planning. 
To train the {\sc parameter selection} agent, we collected a corpus of 1080 publicly available and highly diverse field robot trajectories.%
We evaluated on 37 instances of 3 distinct environments, where our proposed method for learning-based parameter selection results in 
a 9.53\% mean improvement in the cumulative reward  when compared to end-to-end RL approaches and a $3.82\%$ mean improvement in the cumulative reward when compared to a baseline method, while expertly chosen parameters only achieve a $0.3\%$ mean improvement when compared to the same baseline.
We demonstrated our approach in the field showing that it is practical and effective for automatic deployment of IPP.

\printbibliography

\end{document}